%% file: main.tex
\pdfoutput=1

\documentclass[11pt]{article}

\usepackage[]{emnlp2021}

\usepackage{times}
\usepackage{latexsym}

\usepackage[T1]{fontenc}

\usepackage[utf8]{inputenc}

\usepackage{microtype}

\usepackage{url}
\usepackage[T1]{fontenc}

\usepackage{booktabs}
\usepackage{bm}
\usepackage{adjustbox}
\usepackage{multirow}
\usepackage{lipsum}
\usepackage{amsfonts}
\usepackage[inline]{enumitem}
\usepackage{amsmath}
\usepackage{trackchanges}
\addeditor{YJ}

\usepackage[disable]{todonotes}

\newcommand{\hit}{HittER\xspace}

\usepackage{array}
\newcolumntype{H}{>{\setbox0=\hbox\bgroup}c<{\egroup}@{}}

\usepackage{xspace,mfirstuc,tabulary}

\title{Knowledge Graph Embeddings from Hierarchical Transformers}
\title{\hit: Hierarchical Transformers for Knowledge Graph Embeddings}

\author{
 Sanxing Chen\Thanks{Work was done during an internship at Microsoft Bing Ads.} \\
 University of Virginia \\
 {\tt sc3hn@virginia.edu} \\ \And
 Xiaodong Liu, Jianfeng Gao \\
 Microsoft Research \\
 {\tt \{xiaodl,jfgao\}@microsoft.com}
 \AND
 Jian Jiao, Ruofei Zhang \\
 Microsoft Bing Ads \\
 {\tt \{jiajia,bzhang\}@microsoft.com} \\ \And
 Yangfeng Ji \\
 University of Virginia \\
 {\tt yangfeng@virginia.edu}
}

\date{}

\begin{document}
\maketitle
\begin{abstract}
This paper examines the challenging problem of learning representations of entities and relations in a complex multi-relational knowledge graph.
We propose \textbf{\hit}, a \textbf{Hi}erarchical \textbf{T}ransformer model \textbf{t}o jointly learn \textbf{E}ntity-relation composition and \textbf{R}elational contextualization based on a source entity's neighborhood.
Our proposed model consists of two different Transformer blocks: the bottom block extracts features of each entity-relation pair in the local neighborhood of the source entity and the top block aggregates the relational information from outputs of the bottom block.
We further design a masked entity prediction task to balance information from the relational context and the source entity itself.
Experimental results show that \hit{} achieves new state-of-the-art results on multiple link prediction datasets.
We additionally propose a simple approach to integrate \hit{} into BERT and demonstrate its effectiveness on two Freebase factoid question answering datasets.
\end{abstract}

\input{sections/1_introduction.tex}
\input{sections/2_methodology.tex}
\input{sections/3_experiments.tex}
\input{sections/5_related_work.tex}
\input{sections/6_conclusion.tex}

\bibliography{anthology,my}
\bibliographystyle{acl_natbib}


\newpage
\appendix

\section{Embedding Clustering}
\input{tables/word_cluster}
Table~\ref{tab:word_cluster} lists the entity clustering results of first few entities in each dataset, based on our learned entity representations. Clusters in FB15K-237 usually are entities of the same type, such as South/Central American countries, government systems, and American voice actresses.
While clusters in WN18RR are generally looser but still relevant to the topic of the central word.

\section{Factoid QA Experiment Details}
In order to connect our HittER model with BERT, we add a cross-attention module after the self-attention module in each BERT layer.
Following the encoder-decoder attention mechanism in \citet{vaswani2017attention}, we use queries from previous BERT layer and keys and values from the output of a corresponding HittER layer.
The pre-trained BERT (base) and HittER models we use have two differences in terms of hyper-parameter settings, i.e., the number of layers and dimentionality.
Since BERT has 12 layers while HittER only has 6 layers, we connect every two BERT layers to one HittER layer and skip the first two layers in BERT.\footnote{
Among various connection strategies, this strategy gives us the best results in pilot experiments, which also suggests that HittER stores different types of information in its multilayer representations.}
Before attention computation, we increase the dimentionality of HittER's output representations to the number of BERT's using linear transformations.
The dimensionality and number of cross-attention heads are set as the same configuration of the BERT base model we use. 

We finetune all of our question answering (QA) models using a batch size of 16 for 20 epochs.
We use the Adam optimizer~\cite{kingma2014adam} with a learning rate of $5e-6$ for all pretrained weights and a learning rate of $5e-5$ for newly added cross-attention modules.
The learning rate linearly increases from 0 over the first 10\% training steps.

\input{sections/4_discussion.tex}

\end{document}

%% file: sections/1_introduction.tex
\section{Introduction}

Knowledge graphs (KG) are a major form of knowledge bases where knowledge is stored as graph-structured data. Because of their broad applications in various intelligent systems including natural language understanding~\cite{logan-etal-2019-baracks,zhang-etal-2019-ernie,hayashi2020latent} and reasoning~\cite{Riedel-etal-2013-relation,xiong-etal-2017-deeppath,bauer-etal-2018-commonsense,verga-etal-2021-adaptable}, learning representations of knowledge graphs has been studied in a large body of literature.

\begin{figure}[tb]
  \centering
  \includegraphics[width=\linewidth]{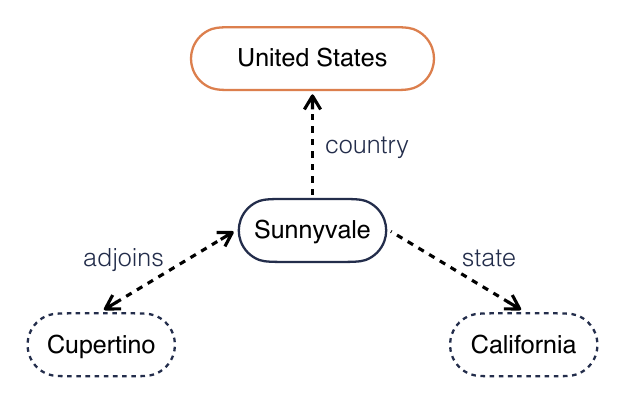}
  \caption{An example subgraph sampled from FB15K-237. Four nodes (entities) are connected by three different types of relations representing facts like \texttt{Sunnyvale} belongs to the state of \texttt{California}.}
  \label{fig:kg}
\end{figure}

To learn high quality representations of knowledge graphs, many researchers adopt the idea of mapping the entities and relations in a knowledge graph to points in a vector space.
These knowledge graph embedding (KGE) methods usually leverage geometric properties in the vector space, such as translation~\cite{bordes2013transe}, bilinear transformations~\cite[DistMult]{yang2014distmult}, or rotation~\cite{sun2018rotate}.
Multi-layer convolutional networks are also used for KGE~\cite[ConvE]{dettmers2018conve}.
Such KGE methods are conceptually simple and can be applied to tasks like factoid question answering~\cite{saxena-etal-2020-improving} and language modeling~\cite{peters-etal-2019-knowledge}.


However, it is rather challenging to encode all of the information about an entity into a single vector.
For example, to infer the missing object in the incomplete triplet \texttt{<Sunnyvale, county, ?>} (Figure~\ref{fig:kg}), traditional KGE methods rely on the geographic information stored in the embedding of \texttt{Sunnyvale}.
While we can read such information from its graph context, e.g., from a neighbor node that represents the state it belongs to (i.e., \texttt{California}).
In this way, we allow the model to store and utilize information about an entity via its relational context.
To implement this process, previous work uses graph neural networks (GNN) or attention-based approaches to learn representations based on both entities and their graph context~\cite{kipf2016semi,bansal-etal-2019-a2n,Vashishth2020Composition}.
However, these methods are usually restricted in expressiveness because of the shallow network architecture they use.\footnote{GNN methods' depth is tied to their receptive fields and thus constrained by over-smoothing issues~\cite{liu2020towards}.}


\begin{figure*}[tb]
  \centering
  \includegraphics[width=\linewidth]{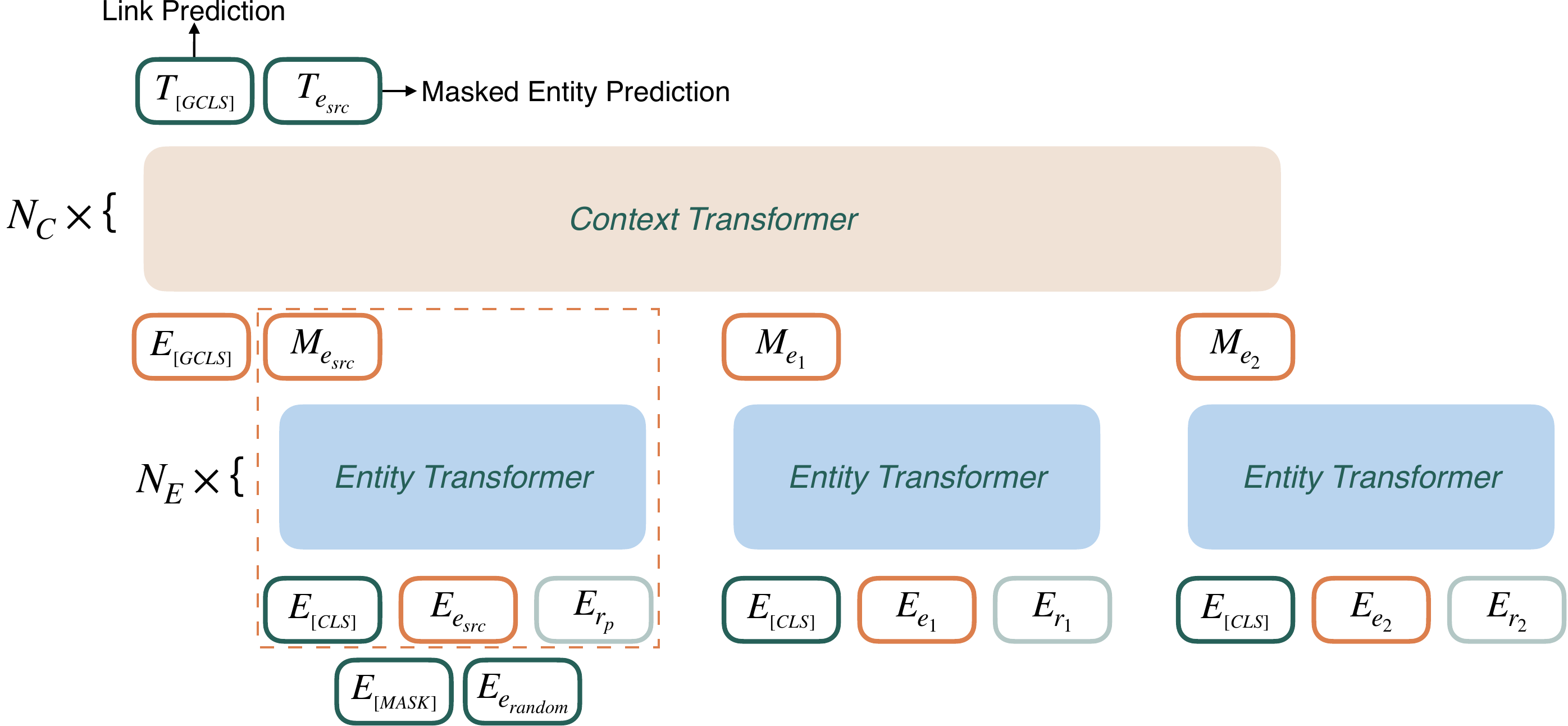}
  \caption{Our model consists of two Transformer blocks organized in a hierarchical fashion. The bottom Transformer block captures the interactions between a entity-relation pair while the top one gathers information from an entity's graph neighborhood. Taking the entity embeddings $E_e$ and the relation embeddings $E_r$ as input, the output embedding $T_{\text{[GCLS]}}$ is used for predicting the target entity. We sometimes mask or replace $E_{e_{\mathit{src}}}$ with $E_{[\mathit{MASK}]}$ or $E_{e_{\mathit{random}}}$. In which case, an additional output embedding $T_{e_{\mathit{src}}}$ can be used to recover the perturbed entity.
  The dashed box indicates a simple context-independent baseline where $M_{e_{\mathit{src}}}$ is directly used for link prediction.}
  \label{fig:model}
\end{figure*}


In this paper, we present \hit, a deep hierarchical Transformer model to learn representations of entities and relations in a knowledge graph jointly by aggregating information from graph neighborhoods.
Although prior work shows Transformers can learn relational knowledge from large amounts of unstructured textual data~\cite{jiang-et-al-2020-how,manning2020emergent},
HittER \emph{explicitly operates over structured inputs using a hierarchical architecture}.
Essentially, HittER consists of two levels of Transformer blocks.
As shown in Figure~\ref{fig:model}, the bottom block provides relation-dependent entity embeddings for the neighborhood around an entity and the top block aggregates information from its graph context.
To ensure HittER work across graphs of different properties, we further design a masked entity prediction task to balance the contextual relational information and information from the training entity itself.


We evaluate the proposed method using the link prediction task, which is one of the canonical tasks in statistical relational learning (SRL).
Link prediction (essentially KG completion) serves as a good proxy to evaluate the effectiveness of learned graph representations, by measuring the ability of a model to generalize relational knowledge stored in training graphs to unseen facts.
Meanwhile, it has an important application to knowledge graph completion given the fact that most of the knowledge graphs are still highly incomplete~\cite{west2014knowledge}.
Our approach achieves new state-of-the-art results on two standard benchmark datasets: FB15K-237~\cite{toutanova-chen-2015-observed} and WN18RR~\cite{dettmers2018conve}.

Unlike the previous shallow KGE methods that cannot be trivially utilized by widely used Transformer-based models for language tasks~\cite{peters-etal-2019-knowledge}, our approach benefits from the unified Transformer architecture and its extensibility.
As a case study, we show how to integrate the learned representations of \hit into pre-trained language models like BERT~\cite{devlin-etal-2019-bert}.
Our experiments demonstrate that \hit{} significantly improves BERT on two Freebase factoid question answering (QA) datasets: FreebaseQA~\cite{jiang-etal-2019-freebaseqa} and WebquestionSP~\cite{yih-etal-2016-value}.

Our experimental code as well as multiple pretrained models are publicly available.\footnote{\url{https://github.com/sanxing-chen/HittER}}


%% file: sections/2_methodology.tex
\section{\hit}
\label{sec:methodology}

We introduce our proposed hierarchical Transformer model (Figure~\ref{fig:model}) in this section. In Section~\ref{sec:link_pred}, we provide the background about how link prediction can be done with a simple Transformer scoring function. We then describe the detailed architecture of our proposed model in Section~\ref{sec:htrme}. Finally, we discuss our strategies to learn balanced contextual representations of an entity in Section~\ref{sec:mem}.

\subsection{Transformers for Link Prediction}
\label{sec:link_pred}

A knowledge graph can be viewed as a set of triplets ($G=\left\{(e_s, r_p, e_o)\right\}$) and each has three items including the subject $e_s\in\mathcal{E}$, the predicate $r_p\in\mathcal{R}$, and the object $e_o\in\mathcal{E}$ to describe a single fact (link) in the knowledge graph. Our model approximates a pointwise scoring function $\psi: \mathcal{E} \times \mathcal{R} \times \mathcal{E} \mapsto \mathbb{R}$ which takes a triplet as input and produces a score reflecting the plausibility of such fact triplet existing in the knowledge graph.
In the task of link prediction, given a triplet with either the subject or the object missing, the goal is to find the missing entity from the set of all entities $\mathcal{E}$.
Without loss of generality, we describe the case where an incomplete triplet $(e_s, r_p)$ is given and we want to predict the object $e_o$.
And vice versa, the subject $e_s$ can be predicted in a similar process, except that a reciprocal predicate $r_{\tilde p}$ will be used to distinguish these two cases~\cite{lacroix2018canonical}.
We call the entity in the incomplete triplet the source entity $e_{\mathit{src}}$ and call the entity we want to predict the target entity $e_{\mathit{tgt}}$.

Link prediction can be done in a straightforward manner with a Transformer encoder~\cite{vaswani2017attention} as the scoring function, depicted inside the dashed box in Figure~\ref{fig:model}.
Our inputs to the Transformer encoder are randomly initialized embeddings of the source entity $e_{\mathit{src}}$, the predicate $r_p$, and a special \texttt{[CLS]} token.
Three different learned type embeddings are directly added to the three token embeddings similar to the input representations of BERT~\cite{devlin-etal-2019-bert}.
Then we use the output embedding corresponding to the \texttt{[CLS]} token ($M_{e_{\mathit{src}}}$) to predict the target entity, which is implemented as follows.
We first compute the plausibility score of the true triplet as the dot-product between $M_{e_{\mathit{src}}}$ and the token embedding of the target entity.
In the same way, we also compute the plausibility scores for all other candidate entities and normalize them using the softmax function. 
Lastly, we use the normalized distribution to get the cross-entropy loss $\mathcal{L}_{\text{LP}}=-\log p(e_{\mathit{tgt}} \mid M_{e_{\mathit{src}}})$ for training.
We will use this model as a simple \emph{context-independent} baseline later in experiments, which is similar to the approach explored in \citet{wang2019coke}.

Although such simple Transformer encoder does a decent work in link prediction tasks, learning knowledge graph embeddings from one triplet at a time ignores the abundant structural information in the graph context. Our model, as described in the following section, also considers the relational neighborhood of the source vertex (entity), which includes all of its adjacent vertices in the graph, denoted as $N_{G}(e_{\mathit{src}}) = \{(e_{\mathit{src}}, r_i, e_i)\}$.\footnote{Our referred neighborhood is slightly different from the formal definition since we only consider edges connecting to the source vertex.}

\subsection{Hierarchical Transformers} \label{sec:htrme}

We propose a hierarchical Transformer model for knowledge graph embeddings (Figure~\ref{fig:model}). The proposed model consists of two blocks of multi-layer bidirectional Transformer encoders. 

We employ the Transformer described in Section~\ref{sec:link_pred} as our bottom Transformer block, called the \emph{entity Transformer}, to learn interactions between an entity and its associated relation type.
Different from the context-independent scenario described in the last section, this entity Transformer is now generalized to also encode information from a relational context.
In specific, there are two cases in our context-dependent scenario:
\begin{enumerate} 
\item We consider the source entity with the predicate in the incomplete triplet as the first entity-relation pair; 
\item We consider an entity from the graph neighborhood of the source entity with the relation type of the edge that connects them.
\end{enumerate}
The bottom block is responsible of packing all useful features from the entity-relation pairs into vector representations to be further used by the top block. Compared with directly feeding all entity-relation pairs to the top block, it helps reduce the run-time of the model by converting two inputs to one.\footnote{This avoids long input sequences for Transformer's $\mathcal{O}(n^2)$ computation.}

The top Transformer block is called the \emph{context Transformer}.
Given the output of the previous entity Transformer and a special \texttt{[GCLS]} embedding, it contextualizes the source entity with relational information from its graph neighborhood.
Similarly, three type embeddings are assigned to the special \texttt{[GCLS]} token embedding, the intermediate source entity embedding, and the other intermediate neighbor entity embeddings.
The cross-entropy loss for link prediction is now changed as follows.

\begin{equation}
\mathcal{L}_{\text{LP}}=-\log p(e_{\mathit{tgt}} \mid T_{[\mathit{GCLS}]})
\end{equation}

The top block does most of the heavy lifting to aggregate contextual information together with the information from the source entity and the predicate, by using structural features extracted from the output vector representations of the bottom block.

\subsection{Balanced Contextualization} \label{sec:mem}
Our hierarchical Transformer model shows a simple way to introduce graph context to link prediction, however, trivially providing contextual information to the model could cause problems.
On one hand, since a source entity often contains high-quality information for link prediction and learning to extract useful information from a broad noisy context requires substantial effort, the model could simply learn to ignore the additional contextual information.
On the other hand, the introduction of rich contextual information could in turn downgrade information from the source entity and contain spurious correlations, which potentially lead to over-fitting based on our observation.
To address these challenges, inspired by the successful Masked Language Modeling pre-training task in BERT, we propose a two-step \emph{Masked Entity Prediction} task (MEP) to balance the utilization of source entity and graph context during contextualization process.

To avoid the first problem, we apply a masking strategy to the source entity of each training example as follows.
During training, we randomly select a proportion of training examples in a batch.
With certain probabilities, we replace the input source entity with a special mask token \texttt{[MASK]}, a random chosen entity, or just leave it unchanged.
The purpose of these perturbations is to introduce extra noise to the information from the source entity, thus forcing the model to learn contextual representations.
The probability of each category is dataset-specific hyper-parameter: for example, we can mask out the source entity more frequently if its graph neighborhood is denser (in which case, the source entity can be easily replaced by the additional contextual information).

In terms of the second problem, we want to promote the model's awareness of the masked entity. Thus we train the model to recover the perturbed source entity based on the additional contextual information.
To do this, we use the output embedding corresponding to the source entity $T_{e_{\mathit{src}}}$ to predict the correct source entity via a classification layer.\footnote{We share the same weight matrix in the input embeddings layer and the linear transformation of this classification layer.}
We can add the cross-entropy classification loss to the previous mentioned link prediction loss as an auxiliary loss, as follows.
\begin{align}
\mathcal{L}_{\text{MEP}}=& -\log p(e_{\mathit{src}} \mid T_{e_{\mathit{src}}}) \\
\mathcal{L} =& \mathcal{L}_{\text{LP}} + \mathcal{L}_{\text{MEP}}
\end{align}
This step is important when solely relying on the contextual clues is insufficient to do link prediction, which means the information from the source entity needs to be emphasized.
And it is otherwise unnecessary when there is high-quality contextual information.
However, the first step of entity masking is always beneficial to the utilization of contextual information according to our observations.
Thus we use dataset-specific configurations to strike a balance between these two sides.

In addition to the MEP task, we implement a uniform neighborhood sampling strategy where only a fraction of the entities in the graph neighborhood will appear in a training example.
This sampling strategy acts like a data augmenter and similar to the edge dropout regularization in graph neural network methods~\cite{Rong2020DropEdge}.
We also have to remove the ground truth target entity from the source entity's neighborhood during training. Otherwise, it will create a dramatic train-test mismatch because the ground truth target entity can always be found from the source entity's neighborhood during training while it can rarely be found during testing. The model will thus learn to naively select an entity from the neighborhood. 

%% file: sections/3_experiments.tex
\section{Link Prediction Experiments}
\label{sec:experiments}

\input{tables/main_results}

We describe our link prediction experiments in this section.
Section~\ref{sec:datasets} introduces two standard benchmark datasets we used.
We then describe our evaluation protocol in Section~\ref{sec:eval_protocal}, and the detailed experimental setup in Section~\ref{sec:exp_setup}. 
Our proposed method are assessed both quantitatively and qualitatively in Section~\ref{sec:results}.
Besides, several ablation studies are presented in Section~\ref{sec:ablations} to demonstrate the importance of balanced contextualization.

\subsection{Datasets}
\label{sec:datasets}
We train and evaluate our proposed method on two standard benchmark datasets FB15K-237~\cite{toutanova-chen-2015-observed} and WN18RR~\cite{dettmers2018conve} for link prediction, following the standard train/test split.\footnote{We intentionally omit the original FB15K and WN18 datasets because of their known flaw in test-leakage~\cite{toutanova-chen-2015-observed}.}
FB15K-237 is a subset sampled from the Freebase~\cite{bollacker2008freebase} with trivial inverse links removed.
It stored facts about topics in movies, actors, awards, etc.
WN18RR is a subset of the WordNet~\cite{miller1995wordnet} which contains structured knowledge of English lexicons.
Statistics of these two datasets are shown in Table~\ref{tab:datasets}.
Notably, WN18RR is much sparser than FB15k-237 which implies it has less structural information in the local neighborhood of an entity.
This will affect our configurations of the masked entity prediction task consequently.

\subsection{Evaluation Protocol}
\label{sec:eval_protocal}
The task of link prediction in a knowledge graph is defined as an entity ranking task.
Essentially, for each test triplet, we remove the subject or the object from it and let the model predict which is the most plausible answer among all possible entities.
After scoring all entity candidates and sorting them by the computed scores, the rank of the ground truth target entity is used to further compute various ranking metrics such as mean reciprocal rank (MRR) and hits@k, $k\in \{1, 3, 10\}$.
We report all of these ranking metrics under the filtered setting proposed in \citet{bordes2013transe} where valid entities except the ground truth target entity are filtered out from the rank list.

\input{tables/datasets}

\subsection{Experimental Setup}
\label{sec:exp_setup}
We implement our proposed method in PyTorch~\cite{pytorch} under the LibKGE framework~\cite{broscheit-etal-2020-libkge}.
To perform a fair comparison with some early baseline methods, we reproduce their results using hyper-parameter configurations from LibKGE.\footnote{These configurations consider many recent training techniques and are found by extensive searches. Thus the results are generally much better then the original reported ones.} All data and evaluation metrics can be found in LibKGE.

Our model consists of a three-layer entity Transformer and a six-layers context Transformer. Each Transformer layer has eight heads. The dimension size of hidden states is 320 across all layers except that we use 1280 dimensions for the position-wise feed-forward networks inside Transformer layers suggested by \citet{vaswani2017attention}.
We set the maximum numbers of uniformly sampled neighbor entities for every example in the FB15K-237 and WN18RR dataset to be 50 and 12 respectively.
Such configurations are intended to ensure most examples (more than 85\% of the cases in each dataset) can have access to its entire local neighborhood during inference.
During training, we further randomly drop 30\% of entities from these fixed-size sets in both datasets.

We train our models using Adamax~\cite{kingma2014adam} with a learning rate of $0.01$ and an L2 weight decay rate of $0.01$.
The learning rate linearly increases from $0$ over the first ten percent of training steps, and linearly decreases through the rest of the steps.
We apply dropout~\cite{srivastava2014dropout} with a probability $p=0.1$ for all layers, except that $p=0.6$ for the embedding layers. 
We apply label smoothing with a rate $0.1$ to prevent the model from being over-confident during training.
We train our models using a batch size of 512 for at most 500 epochs and employ early stopping based on MRR in the validation set.

When training our model with the masked entity prediction task, we use the following dataset-specific configurations based on validation MRR in few early trials:
\begin{itemize}
    \item \textbf{WN18RR}: 50\% of examples are subjected to this task. Among them, 60\% of examples are masked out, the rest are split in a 3:7 ratio for replaced and unchanged ones.
    \item \textbf{FB15K-237}: 50\% of examples in a batch are masked out. No replaced or unchanged ones. We do not include the auxiliary loss.
\end{itemize}

Training our full models takes 7 hours (WN18RR) and 37 hours (FB15K-237) on a NVIDIA Tesla V100 GPU.

\subsection{Experimental Results}
\label{sec:results}

Table~\ref{tab:main} shows that the results of \hit{} compared with baseline methods including some early methods and previous SOTA methods.\footnote{We do not compare with huge models that employ excessive embeddings size~\cite{lacroix2018canonical}.} We outperform all previous work by a substantial margin across nearly all the metrics.
Comparing to some previous methods which target some observed patterns of specific datasets, our proposed method is more general and is able to give more consistent improvements over the two standard datasets.
For instance, the previous SOTA in WN18RR, RotH explicitly captures the hierarchical and logical patterns by hyperbolic embeddings.
Comparing to it, our model performs better especially in the FB15K-237 dataset which has a more diverse set of relation types.
On the other hand, our models have comparable numbers of parameters to baseline methods, since entity embeddings contribute to the majority of the parameters.


\subsection{Ablation Studies} \label{sec:ablations}
\input{tables/ablations}

To show the contributions of adding graph context and balanced contextualization, we compare results of three different settings (Table~\ref{tab:ablations}), i.e., \hit{} with no context (the context-independent Transformer described in Section~\ref{sec:link_pred}), contextualized \hit{} without balancing techniques proposed in Section~\ref{sec:mem}, and our full model.
We find that directly adding in contextual information does not benefit the model (``Unbalanced''), while balanced contextualization generates significantly superior results in terms of MRR on both datasets, especially for the WN18RR dataset, which has a sparser and noisier graph structure.

Breaking down the model's performance by relation types in WN18RR, Table~\ref{tab:wnrr_type} shows that incorporating contextual information brings us substantial improvements on two major relation types, namely the \emph{hypernym} and the \emph{member meronym} relations, which both include many examples belong to the challenging one-to-many relation categories defined in \citet{bordes2013transe}.

\input{tables/wnrr_by_type}

\begin{figure}[t]
  \centering
  \includegraphics[width=\linewidth]{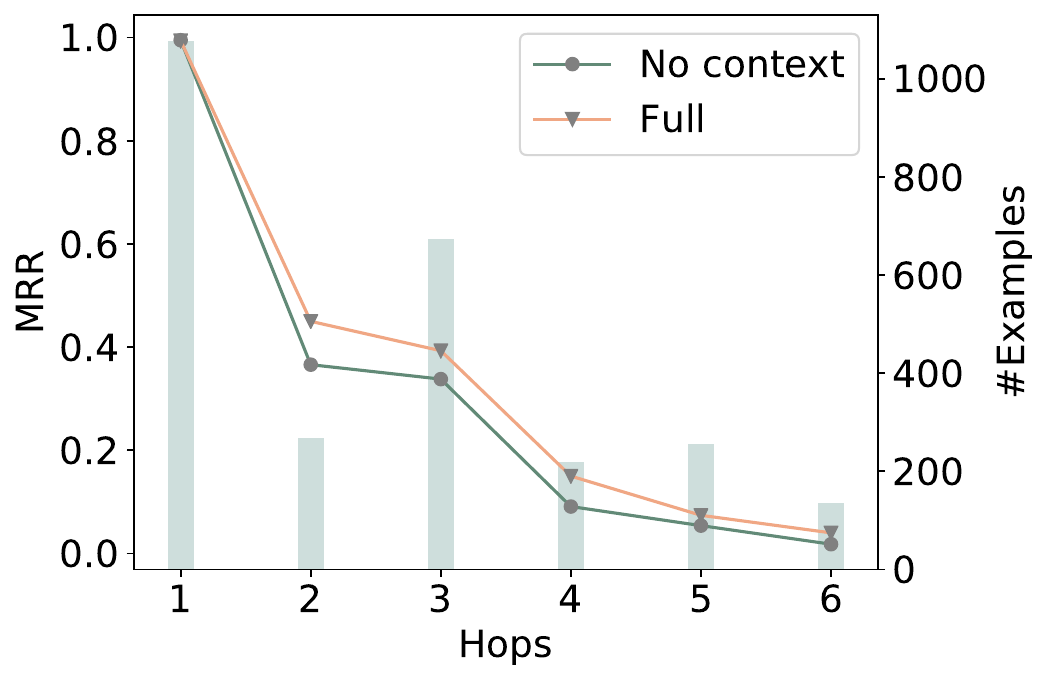}
  \caption{Dev mean reciprocal rank (MRR) in the WN18RR dataset grouped by the number of hops. The bar chart shows the number of examples in each group.}
  \label{fig:hops}
\end{figure}

Inferring the relationship between two entities can be viewed as a process of aggregating information from the graph paths between them~\cite{teru2019inductive}.
To gain further understanding of what the role the contextual information play from this aspect, we group examples in the development set of WN18RR by the number of hops (\textit{i.e.}, the shortest path length in the undirected training graph) between the subject and the object in each example (Figure~\ref{fig:hops}). From the results, we can see that the MRR metric of each group decreases by the number of hops of the examples.
This matches our intuition that aggregating information from longer graph paths is generally harder and such information is more unlikely to be meaningful.
Comparing models with and without the contextual information, the contextual model performs much better in groups of multiple hops ranging from two to four.
The improvement also shrinks as the number of hops increases.

\begin{figure}[t]
  \centering
  \includegraphics[width=\linewidth]{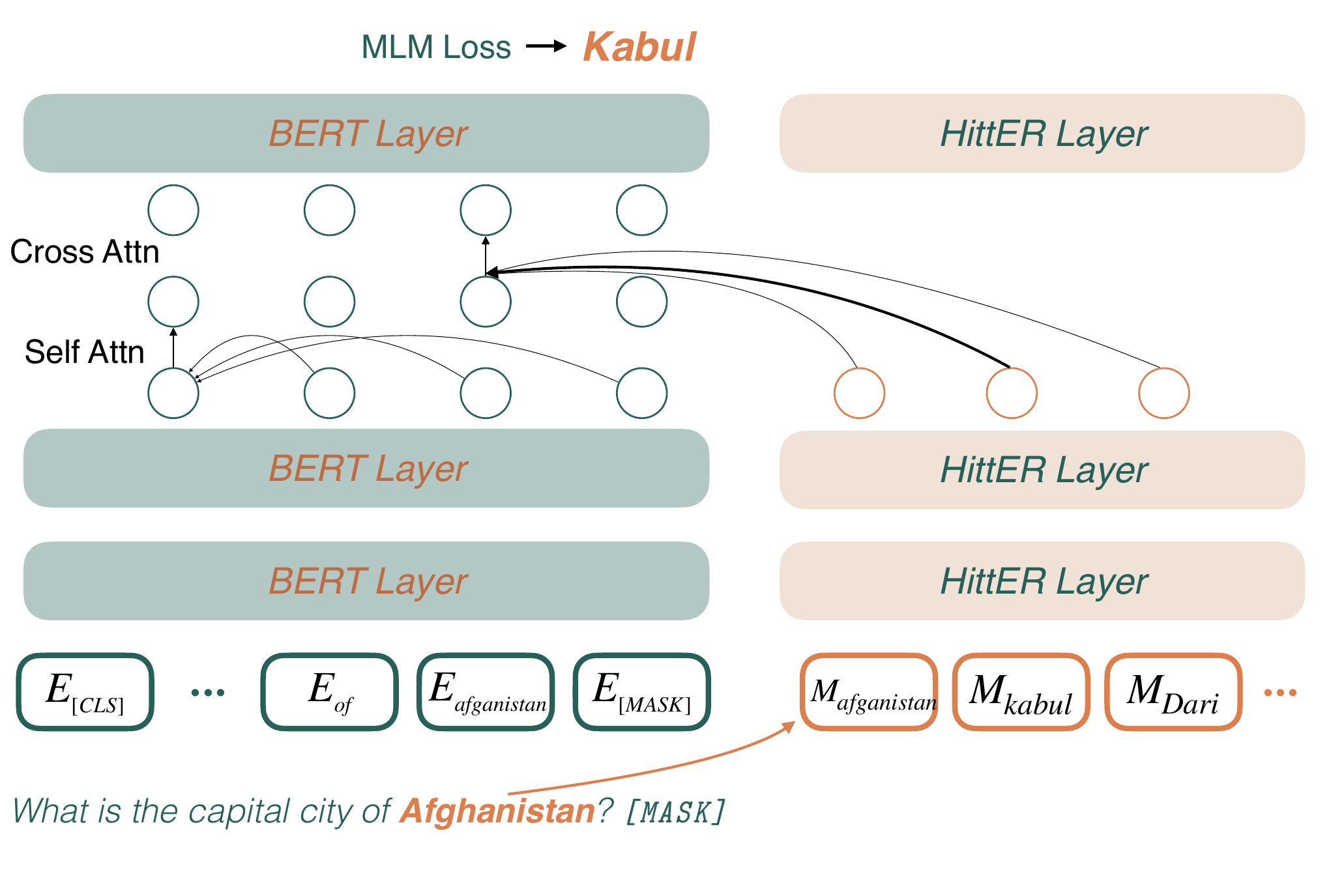}
  \caption{Combining HittER and BERT for factoid QA. Each BERT layer is connected to a layer of HittER's context Transformer via a cross-attention module. We jointly fine-tune the combined model to predict the masked entity name in the input question.}
  \label{fig:hitbert}
\end{figure}

\section{Factoid QA Experiments}
\label{sec:qa-experimens}

In addition to HittER's superior intrinsic evaluation results, in this section, we conduct a case study on the factoid question answering (QA) task to demonstrate HittER's potential to enhance popular pre-trained Transformer-based language models' performance on knowledge-intensive tasks.

As a Transformer-based model, HittER enables us to integrate its multilayer knowledge representation into other Transformer models (BERT in our case) using the multi-head attention mechanism.
In each BERT layer, after the original self-attention module we add a cross-attention module where the queries come from the previous BERT layer while the keys and values come from the output of a corresponding HittER layer~\cite{vaswani2017attention}, so that HittER's knowledge information can flow into BERT (Figure~\ref{fig:hitbert}).

We perform experiments on two factoid QA datasets: FreebaseQA~\cite{jiang-etal-2019-freebaseqa} and WebQuestionSP~\cite{yih-etal-2016-value}, both pertaining to facts on Freebase.
Each question in the two datasets is labeled with a context entity and an inferred relation between the context entity and the answer entity, which we use for preparing the entity and relation inputs for HittER.
To better exploit the knowledge in BERT, we follow its pretraining task to create a word-based QA setting, where factoid questions are converted to cloze questions by appending the special \texttt{[MASK]} tokens to the end.
Both models are trained to recover these \texttt{[MASK]} tokens to the original words.\footnote{This is different from the entity-based QA setting. To simplify the modeling architecture, we also make the number of tokens known to all models.}
We use the BERT-base model~\cite{devlin-etal-2019-bert} and our best performing HittER model pre-trained on the FB15K-237 dataset.
Since FB15K-237 only covers a small portion of Freebase, most questions in the two QA datasets are not related to the knowledge from the FB15K-237 dataset, in which case the input entities for HittER cannot be provided.
Thus we also report results under a \emph{filtered} setting, i.e., a subset retaining only examples whose context entity and answer entity both exist on the FB15K-237 dataset.

\input{tables/qa_datasets}
\input{tables/qa_results}

Our experimental results in Table~\ref{tab:qa-results} show that HittER's representation significantly enhances BERT's question answering ability, especially when the questions are related to entities in the knowledge graph used to train HittER.
We include more details of the experiments in the Appendix.


%% file: tables/main_results.tex
\begin{table*}[tb]
\centering
\setlength{\tabcolsep}{3pt}
\begin{adjustbox}{max width=\textwidth}
\begin{tabular}{lrrHrrrrrHrrr}
\toprule
\multirow{3}{*}{Model}  & \multicolumn{6}{c}{FB15K-237} & \multicolumn{6}{c}{WN18RR} \\\cmidrule(r{4pt}){2-7} \cmidrule(l){8-13}
      & \multirow{2}{*}{\#Params}  & \multirow{2}{*}{MRR$\uparrow$} & \multirow{2}{*}{MR$\downarrow$} & \multicolumn{3}{c}{Hits$\uparrow$}  & \multirow{2}{*}{\#Params} & \multirow{2}{*}{MRR$\uparrow$} & \multirow{2}{*}{MR$\downarrow$} & \multicolumn{3}{c}{Hits$\uparrow$} \\\cmidrule(r{4pt}){5-7}\cmidrule(l){11-13}
      & & & & @1 & @3 & @10 & & & & @1 & @3 & @10 \\\midrule
RESCAL~\cite{nickel2011rescal}   &  6M & .356 & 192 & .266 & .390 & .535 &  6M & .467 & 5408 & .439 & .478 & .516 \\
TransE~\cite{bordes2013transe}   &  2M & .310 & 199 & .218 & .345 & .495 &  21M & .232 & 1706 & .061 & .366 & .522 \\
DistMult~\cite{yang2014distmult} &  4M & .342 & 177 & .249 & .378 & .531 & 21M & .451 & 4390 & .414 & .466 & .523 \\
ComplEx~\cite{trouillon2016complexe}  &  4M & .343 & 176 & .250 & .377 & .532 &  5M & .479 & 3817 & .441 & .495 & .552 \\
ConvE~\cite{dettmers2018conve}    &  9M & .338 & 176 & .247 & .372 & .521 & 36M & .439 & 4789 & .409 & .452 & .499 \\
RotatE~\cite{sun2018rotate}   &  15M  & .338 & 177 & .241 & .375 & .533 &  20M  & .476 & 3340 & .428 & .492 & .571 \\
CoKE~\cite{wang2019coke}  &  10M  & .364 & - & .272 & .400 & .549 &  17M  & .484 & - & .450 & .496 & .553 \\
TuckER~\cite{balazevic-etal-2019-tucker}  &  -  & .358 & - & .266 & .394 & .544 &  -  & .470 & - & .443 & .482 & .526 \\
CompGCN~\cite{Vashishth2020Composition}  &  -  & .355 & 197 & .264 & .390 & .535 &  -  & .479 & 3533 & .443 & .494 & .546 \\
RotH~\cite{chami-etal-2020-low}    &  8M  & .344 & -   & .246 & .380 & .535 &  21M  & .496 & -    & .449 & .514 & \textbf{.586} \\\midrule
\hit     & 16M & \textbf{.373} & \textbf{158} & \textbf{.279} & \textbf{.409} & \textbf{.558} & 24M & \textbf{.503} & \textbf{2268} & \textbf{.462} & \textbf{.516} & .584 \\
\bottomrule
\end{tabular}
\end{adjustbox}
\caption{Comparison between the proposed method and baseline methods. Results of RotatE, CoKE, TuckER, CompGCN, and RotH are taken from their original papers. Numbers in \textbf{bold} represent the best results.}
\label{tab:main}
\end{table*}

%% file: tables/datasets.tex
\begin{table}[t]
\centering
\begin{tabular}{lrr}
\toprule
Dataset       & FB15K-237 & WN18RR \\\midrule
\#Entities    & 14,541    & 40,943 \\
\#Relations   & 237       & 11     \\
\#Triples     & 310,116   & 93,003 \\
\#Avg. degree & 42.7      & 4.5   \\\bottomrule
\end{tabular}
\caption{Dataset statistics. The WN18RR dataset is significantly sparser than the FB15K-237 dataset.}
\label{tab:datasets}
\end{table}

%% file: tables/ablations.tex
\begin{table}[t]
\setlength{\tabcolsep}{3pt}
\begin{adjustbox}{max width=\linewidth}
\begin{tabular}{lrrrr}
\toprule
\multirow{2}{*}{\normalsize{Contextualization}} & \multicolumn{2}{c}{FB15K-237} & \multicolumn{2}{c}{WN18RR} \\\cmidrule(r{4pt}){2-3} \cmidrule(l){4-5}
                       & MRR           & H@10          & MRR          & H@10        \\\midrule
Balanced            & \textbf{37.5\textsubscript{(.1)}}          & \textbf{56.1\textsubscript{(.1)}}        & \textbf{50.0\textsubscript{(.4)}}        & \textbf{58.2\textsubscript{(.4)}}      \\
Unbalanced                 & 36.7\textsubscript{(.2)}         & 55.4\textsubscript{(.4)}         & 47.5\textsubscript{(.1)}        & 55.4\textsubscript{(.2)}       \\
None             & 37.3\textsubscript{(.1)}         & 56.1\textsubscript{(.1)}         & 47.3\textsubscript{(.6)}        & 53.8\textsubscript{(.7)}     \\\bottomrule
\end{tabular}
\end{adjustbox}
\caption{Results of models with different contextualization techniques on dev sets. We report average scores and standard deviation from five random runs.}
\label{tab:ablations}
\end{table}

%% file: tables/wnrr_by_type.tex
\begin{table}[tb]
\centering
\setlength{\tabcolsep}{3pt}
\begin{adjustbox}{max width=\linewidth}
\begin{tabular}{lrrrr}
\toprule
Relation Name    & Count & No ctx & Full & Gain \\\midrule
 hypernym & 1174 & .144 & .181 & 26\%\\
 derivationally related form & 1078 & .947 & .947 & 0\%\\
 member meronym & 273 & .237 & .316 & 33\%\\
 has part & 154 & .200 & .235 & 18\%\\
 instance hypernym & 107 & .302 & .330 & 9\%\\
 synset domain topic of & 105 & .350 & .413 & 18\%\\
 verb group & 43 & .930 & .931 & 0\%\\
 also see & 41 & .585 & .595 & 2\%\\
 member of domain region & 34 & .201 & .259 & 29\%\\
 member of domain usage & 22 & .373 & .441 & 18\%\\
 similar to & 3 & 1 & 1 & 0\%\\\bottomrule
\end{tabular}
\end{adjustbox}
\caption{Dev MRR and relative improvement percentage of our proposed method with or without the \emph{context Transformer} respect to each relation in the WN18RR dataset.}
\label{tab:wnrr_type}
\end{table}

%% file: tables/qa_datasets.tex
\begin{table}[t]
\centering
\begin{adjustbox}{max width=\linewidth}
\begin{tabular}{lrrrr}
\toprule
\multirow{2}{*}{} & \multicolumn{2}{c}{FreebaseQA} & \multicolumn{2}{c}{WebQuestionSP} \\\cmidrule(r{4pt}){2-3} \cmidrule(l){4-5}
                  & Full   & Filtered  & Full    & Filtered    \\\midrule
Train             & 20358          & 3713          & 3098            & 850             \\
Test              & 3996           & 755           & 1639            & 484            \\\bottomrule
\end{tabular}
\end{adjustbox}
\caption{Number of examples in two Freebase question answering datasets.}
\label{tab:qa-dataset-statistics}
\end{table}

%% file: tables/qa_results.tex
\begin{table}[t]
\centering
\begin{adjustbox}{max width=\linewidth}
\begin{tabular}{lrrrr}
\toprule
\multirow{2}{*}{Model} & \multicolumn{2}{c}{FreebaseQA} & \multicolumn{2}{c}{WebQuestionSP} \\\cmidrule(r{4pt}){2-3} \cmidrule(l){4-5}
                  & Full   & Filtered  & Full    & Filtered   \\\midrule
BERT              & 19.8\textsubscript{(.1)} & 30.8\textsubscript{(.1)}        & 23.2\textsubscript{(.3)} & 46.5\textsubscript{(.4)} \\
+HittER           & \textbf{21.2\textsubscript{(.2)}}          & \textbf{37.1\textsubscript{(.6)}}        & \textbf{27.1\textsubscript{(.2)}}          & \textbf{51.0\textsubscript{(.7)}}         \\\bottomrule
\end{tabular}
\end{adjustbox}
\caption{QA accuracy of combining HittER and BERT in two Freebase-based question answering datasets. We report average scores and standard deviation from five random runs.}
\label{tab:qa-results}
\end{table}

%% file: sections/5_related_work.tex
\section{Related Work}
\label{sec:related}

KGE methods have been extensively studied in many diverse directions. Our scope here is limited to methods that purely rely on entities and relations, without access to other external resources.

\subsection{Triple-based Methods}

Most of the previous work focuses on exploiting explicit geometric properties in the embedding space to capture different relations between entities.
Early work uses translational distance-based scoring functions defined on top of entity and relation embeddings~\cite{bordes2013transe,wang2014transh,lin2015tranr,ji-etal-2015-knowledge}.

Another line of work uses tensor factorization methods to match entities semantically. Starting from simple bi-linear transformations in the euclidean space~\cite{nickel2011rescal,yang2014distmult}, numerous complicated transformations in various spaces have been hence proposed~\cite{trouillon2016complexe,ebisu2018toruse,sun2018rotate,zhang2019quaternion,chami-etal-2020-low,tang-etal-2020-orthogonal,chao-etal-2021-pairre}. Such methods effectively capture the intuition from observation of data but suffer from unobserved geometric properties and are generally limited in expressiveness.

In light of recent advances in deep learning, many more powerful neural network modules such as Convolutional Neural Networks~\cite{dettmers2018conve}, Capsule Networks~\cite{nguyen-etal-2019-capsule}, and Transformers~\cite{wang2019coke} are also introduced to capture the interaction between entity and relation embeddings.
These methods produce rich representations and better performance on predicting missing links in knowledge graphs.
However, they are restricted by the amount of information that can be encoded in a single node embedding and the great effort to memorize local connectivity patterns. 

\subsection{Context-aware Methods}

Various forms of graph contexts have been proven effective in recent work on neural networks operating in graphs under the message passing framework~\cite{bruna2013spectral,defferrard2016convolutional,kipf2016semi}.
\citet[R-GCN]{schlichtkrull2018rgcn} adapt the Graph Convolutional Networks to realistic knowledge graphs which are characterized by their highly multi-relational nature.
\citet{teru2019inductive} incorporate an edge attention mechanism to R-GCN, showing that the relational path between two entities in a knowledge graph contains valuable information about their relations in an inductive learning setting.
\citet{Vashishth2020Composition} explore the idea of using existing knowledge graph embedding methods to improve the entity-relation composition in various Graph Convolutional Network-based methods.
\citet{bansal-etal-2019-a2n} borrow the idea from Graph Attention Networks~\cite{velivckovic2018graph}, using a bi-linear attention mechanism to selectively gather useful information from neighbor entities. Different from their simple single-layer attention formulation, we use the advanced Transformer to capture both the entity-relation and entity-context interactions.
\citet{nathani-etal-2019-learning} also propose an attention-based feature embedding to capture multi-hop neighbor information, but unfortunately, their reported results have been proven to be unreliable in a recent re-evaluation~\cite{sun-etal-2020-evaluation}.

%% file: sections/6_conclusion.tex
\section{Conclusion and Future Work}
\label{sec:conclusion}
In this work, we proposed \hit, a novel Transformer-based model with effective training strategies for learning knowledge graph embeddings in complex multi-relational graphs. We show that with contextual information from a local neighborhood, our proposed method outperforms all previous approaches in long-standing link prediction tasks, achieving new SOTA results on FB15K-237 and WN18RR.
Moreover, we show that the knowledge representation learned by HittER can be effectively utilized by a Transformer-based language model BERT to answer factoid questions.

It is worth mentioning that our proposed balanced contextualization is also applicable to other context-aware KGE methods such as GNN-based approaches.
Future work can also apply \hit{} to other graph representation learning tasks besides link prediction.
Currently, our proposed \hit{} model performs well while only aggregating contextual information from a local graph neighborhood.
It would be interesting to extend it with a broader graph context to obtain potential improvements.

Today, the Transformer has become the de facto modeling architecture in natural language processing.
As experimental results on factoid question answering tasks showcasing \hit{}'s great potential to be integrated into common Transformer-based model and generate substantial gains in performance, we intend to explore training \hit{} on large-scale knowledge graphs, so that more NLP models would benefit from \hit{} in various knowledge-intensive tasks.

\section*{Acknowledgments}
\label{sec:ack}
We thank Hao Cheng, Hoifung Poon, Xuan Zhang, Yu Bai, Aidan San, colleagues from Microsoft Bing Ads team and Microsoft Research, and
the anonymous reviewers for their valuable discussions and comments.

%% file: tables/word_cluster.tex
\begin{table*}[tb]
\centering
\begin{tabular}{p{0.2\linewidth}p{0.75\linewidth}}
\toprule
Entity    & Top 5 Neighbors \\\midrule
Dominican Republic & Costa Rica, Ecuador, Puerto Rico, Colombia, El Salvador  \\
Republic & Presidential system, Unitary state, Democracy, Parliamentary system, Constitutional monarchy  \\
MMPR & Power Rangers, Sonic X, Ben 10, Star Trek: Enterprise, Code Geass  \\
Wendee Lee & Liam O'Brien, Michelle Ruff, Hilary Haag, Chris Patton, Kari Wahlgren  \\
Drama & Thriller, Romance Film, Mystery, Adventure Film, LGBT  \\
\midrule
Land reform& Pronunciamento, Premium, Protest march, Reform, Birth-control reformer\\
Reform& Reform, Land reform, Optimization, Self-reformation, Enrichment\\
Cover& Surface, Spread over, Bind, Supply, Strengthen\\
Covering& Sheet, Consumer goods, Flap, Floor covering, Coating\\
Phytology& Paleobiology, Zoology, Kingdom fungi, Plant life, Paleozoology\\
\bottomrule
\end{tabular}
\caption{Nearest neighbors of first five entities in FB15K-237 and WN18RR based on the cosine similarity between learned entity embeddings from our proposed method.}
\label{tab:word_cluster}
\end{table*}

%% file: sections/4_discussion.tex
\section{Discussion}
\label{sec:discussion}

\subsection{Right Context for Link Prediction}

Structural information of knowledge graphs can come from multiple forms, such as graph paths, sub-graphs, and the local neighborhood that we used in this work. In addition, these context forms can be represented in terms of the relation type, the entity, or both of them. 

In this work, we show that a simple local neighborhood is sufficient to greatly improve a link prediction model.
In early experiments in the FB15K-237 dataset, we actually observe that masking out the source entity all the time does not harm the model performance much.
This implies that the contextual information in a dense knowledge graph like FB15K-237 is rich enough to replace the source entity in the link prediction task.

Recently, \citet{wang2020entity} argue that graph paths and local neighborhood should be jointly considered when only the relation types is used (throwing out entities).
Although some recent work has made a first step towards utilizing graph paths for knowledge graph embeddings~\cite{wang2019coke,wang2020dolores}, there is still no clear evidence of its effectiveness.

\subsection{Limitations of the \emph{1vsAll} Scoring}
Recall that \hit{} learns a representation for an incomplete triplet $(e_s, r_p)$ and then computes the dot-product between it and all the candidate target entity embeddings. This two-way scoring paradigm, which is often termed \textit{1vsAll} scoring, supports fast training and inference when the interactions between the source entity and the predicate are captured by some computation-intensive operations (\textit{i.e.}, the computations of Transformers in our case), but unfortunately loses three-way interactions.
We intentionally choose 1vsAll scoring for two reasons.
On one hand, 1vsAll together with cross-entropy training has shown a consistent improvement over other alternative training configurations empirically~\cite{Ruffinelli2020You}.
On the other hand, it ensures a reasonable speed for the inference stage where the 1vsAll scoring is necessary.

Admittedly, early interactions between the source entity and the target entity can provide valuable information to inform the representation learning of the incomplete triplet $(e_s, r_p)$.
For instance, we find that a simple bilinear formulation of the source entity embeddings and the target entity embeddings can be trained to reflect the distance (measured by the number of hops) between the source entity and the target entity in the graph.
We leave the question of how to effectively and efficiently incorporate such early fusion for future work.